\documentclass[letterpaper, 10 pt, conference]{ieeeconf}  

\IEEEoverridecommandlockouts                              

\usepackage{graphicx} 
\usepackage{amsmath} 
\usepackage{amssymb}  
\usepackage{algorithm}
\usepackage{algpseudocode}
\usepackage{xcolor}

\title{\LARGE \bf
Towards Safe and Efficient Through-the-Canopy Autonomous Fruit Counting with UAVs
}

\author{Teaya Yang, Roman Ibrahimov and Mark W. Mueller
\thanks{The authors are with the High Performance Robotics Lab, Department of Mechanical Engineering, University of California, Berkeley, CA 94720, USA.}}

\begin{document}

\maketitle
\thispagestyle{empty}
\pagestyle{empty}

\begin{abstract}
We present an autonomous aerial system for safe and efficient through-the-canopy fruit counting. Aerial robot applications in large-scale orchards face significant challenges due to the complexity of fine-tuning flight paths based on orchard layouts, canopy density, and plant variability. Through-the-canopy navigation is crucial for minimizing occlusion by leaves and branches but is more challenging due to the complex and dense environment compared to traditional over-the-canopy flights. Our system addresses these challenges by integrating: i) a high-fidelity simulation framework for global path planning, ii) a low-cost autonomy stack for canopy-level navigation and data collection, and iii) a robust workflow for fruit detection and counting using RGB images. We validate our approach through fruit counting with canopy-level aerial images and by demonstrating the autonomous navigation capabilities of our experimental vehicle.

\end{abstract}

\section{Introduction}

Fruit counting in orchards helps farmers make management decisions, such as harvest scheduling, labor allocation, and storage strategies \cite{zhang2021management}. With recent advances in precision agriculture and remote sensing, robotic systems have demonstrated great success in automating the counting process, significantly reducing labor cost and enabling frequent crop inspections. 

In particular, fruit counting using sequences of RGB images collected by autonomous systems through computer vision algorithms \cite{farjon2023deepsurvey} has enabled robots equipped with low-cost sensors to perform yield estimation and prediction tasks. 
For instance, \cite{maldonado2016green} presents a method to reliably detect green fruits using RGB images, in contrast to previous techniques that relied on expensive multispectral sensors \cite{aleixos2002multispectral}. Robust fruit detection of holly fruits, which are small and occur in clusters, is demonstrated in \cite{zhang2022holly}, further highlighting the capability of modern algorithms to extract valuable information using only colored images. The You Only Look Once (YOLO) algorithm \cite{redmon2016yolo} has become a prominent method in fruit detection tasks \cite{zhang2022holly}-\cite{parico2021yolo3}, and its successive versions continue to lead advancements in this field. Building upon these successes, we incorporate YOLO as the primary detection tool for our proposed autonomous counting method.

\begin{figure}
    \centering
    \includegraphics[width=\linewidth]{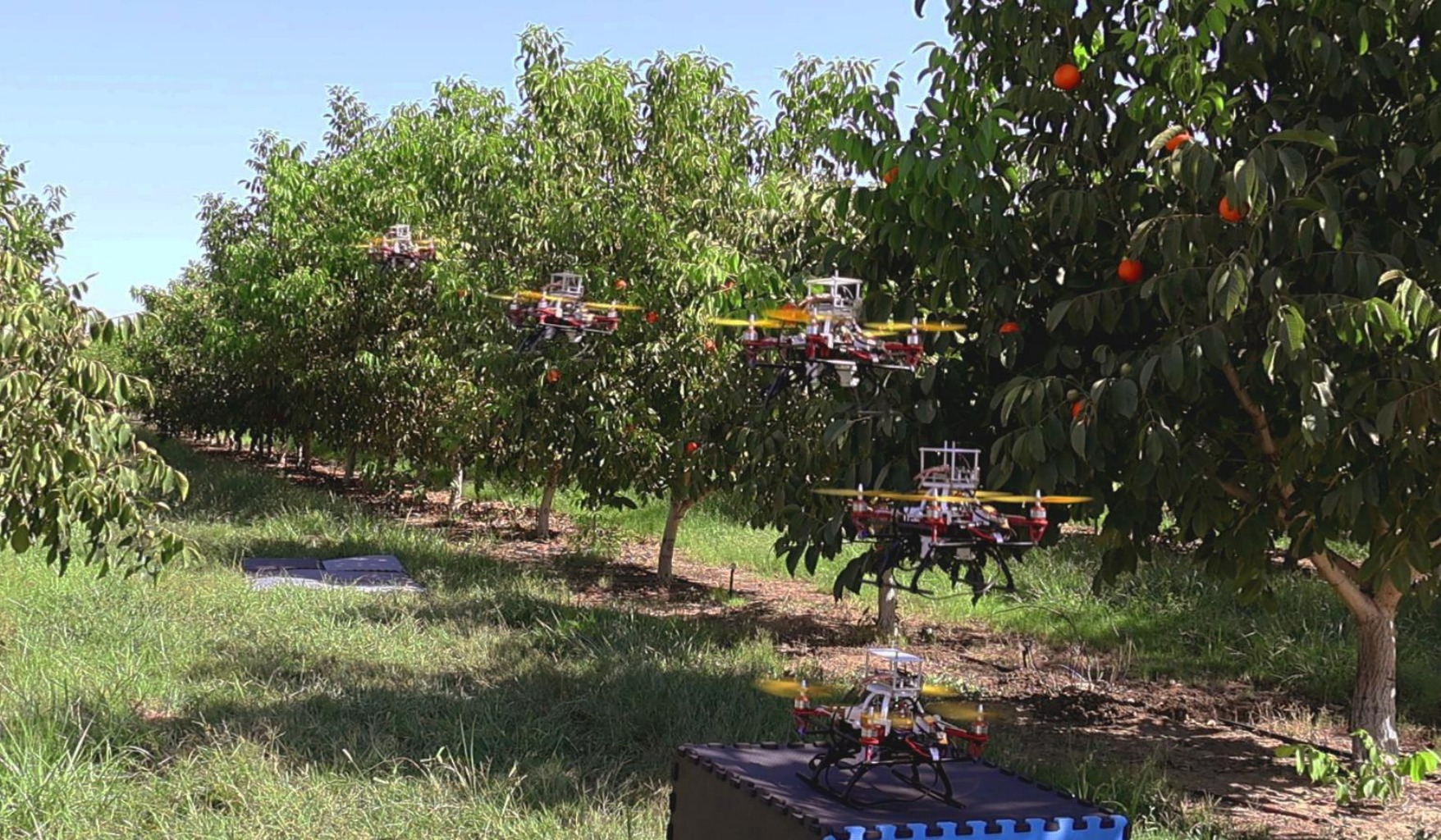}
    \caption{Experimental vehicle flying autonomously at canopy level while collecting visual data using a RGB camera. This figure is a composite image showing multiple states of the vehicle during flight.}
    \label{Fig:FlightSequence}
\end{figure}

Detection alone, however, is not sufficient for providing reliable fruit counting results using visual data. The major challenge in fruit counting with image sequence is the risk of double counting, especially when the same fruit appears multiple times across different frames. In \cite{liu2019monocular}, a robust fruit counting framework that integrates detection, tracking, and structure from motion is outlined. Detected fruits are used as features and tracked using the Hungarian Algorithm \cite{kuhn1955hungarian}, with the results serving as raw inputs for the structure-from-motion step using COLMAP \cite{schoenberger2016colmap,schoenberger2016colmap2}. We adopt this method in our work, tailoring it to data collected from canopy-level flight, and incorporate up-to-date tracking techniques. Specifically, we use ByteTrack \cite{zhang2022bytetrack} in place of the Hungarian Algorithm to better maintain fruit tracking, particularly in cases of low-confidence detections. This approach is crucial when image quality is reduced due to factors such as strong sunlight or motion blur, conditions that are common in outdoor environments.

\begin{figure*}[h]
    \centering
    \includegraphics[width=\textwidth]{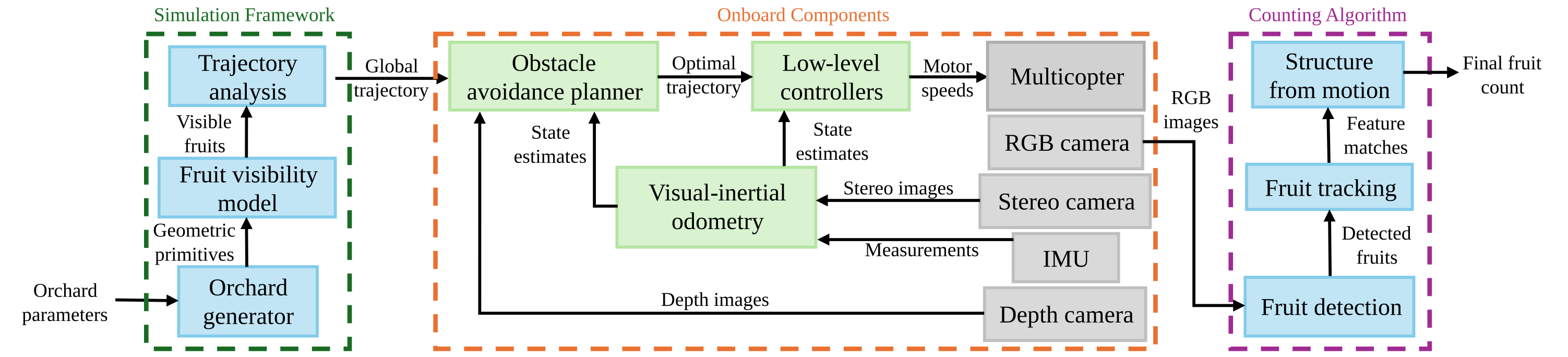}
    \caption{Block diagram of the proposed autonomous system, consisting of three main components: a simulation framework that leverages specified orchard parameters (such as average tree height, tree spacing, and plant type) for global planning; onboard autonomy components that ensure safe and efficient data collection; and a post-flight counting algorithm that processes the collected RGB images, producing the final fruit count.}
    \label{Fig:BlockDiagram}
\end{figure*}

Despite the successes of computer vision techniques in image processing, reliable data collection using autonomous robotic systems remains a challenge. Previous efforts to ensure the safety of robots include using UAVs to fly above tree canopies \cite{mokrane2019overtree, apolo2020overtree}, enabling rapid data collection thanks to their agility and maneuverability. An alternative solution involves operating ground vehicles \cite{mokrane2020groundcoverage} between tree rows, which has shown promise in counting large fruits in orchards with wall-like canopies, where ground-level data is sufficient to capture most fruits. Nevertheless, both methods miss critical vantage points that are essential for detecting fruits in trees with dense foliage. Flying at canopy level offers a better solution, as illustrated in Figure~\ref{Fig:three_traj_comp}, which compares simulated fruit counting results in a walnut orchard. While some works focusing on image processing collected data using manually controlled canopy-level flights \cite{zheng2023citrus}, \cite{stefas2019orcharduav} addressed this challenge by using stereo images for obstacle avoidance through onboard local planning. Our approach also ensures safe autonomous navigation using minimal onboard sensors and a memoryless local planner, while providing a more integrated solution by incorporating global planning and a complete fruit counting workflow.

In the context of through-the-canopy flight, global planning methods are extremely limited. While \cite{stefas2019orcharduav} addresses the challenges of safety and navigation, their approach to global planning assumes fruit trees have wall-like properties and applies coverage-path planning techniques commonly used in structural inspection \cite{galceran2013structralsurvey}. These methods are inadequate for orchards with dense foliage, where fruits within the camera’s field of view may be easily occluded by leaves and branches. Effective path planning must account for these occlusions to optimize fruit visibility. Previous efforts that consider these spatial relationships, such as  \cite{roy2017activeview} and \cite{marangoz2022shape}, focus on inspecting individual fruit clusters from a close distance, making these methods unsuitable for orchard-scale planning, where hundreds of fruits may appear in a single image. Moreover, plant science research like \cite{wang2023distribution} has shown that fruit distribution varies across different levels of the tree canopy, making it essential to account for both local and global fruit distributions when planning paths at the orchard level. Our proposed method provides a more comprehensive solution to this planning problem by incorporating high-fidelity fruit distribution and occlusion models, thereby optimizing flight paths for efficient and large-scale data collection.

Our proposed system achieves autonomous fruit counting in orchard environments through three main steps: global planning based on high-fidelity modeling, data collection using an autonomous vehicle, and fruit detection and counting from onboard RGB images. 
As illustrated in Figure~\ref{Fig:BlockDiagram}, the system begins with a simulation framework, described in Section~\ref{Sec:Sim}, which uses high-fidelity tree models with realistic architectures to compute visible fruit counts from predefined flight paths. The next step, outlined in Section~\ref{Sec:Auto}, involves data collection with an experimental vehicle equipped with minimal sensors. This vehicle ensures safe, autonomous operation through vision-based state estimation and depth-image-based obstacle avoidance. Finally, Section~\ref{Sec:Counting} presents the fruit detection and counting workflow incorporating state-of-the-art tools for detection, tracking, and structure from motion to process collected RGB images.

\begin{figure}[t]
    \centering
    \includegraphics[width=\linewidth]{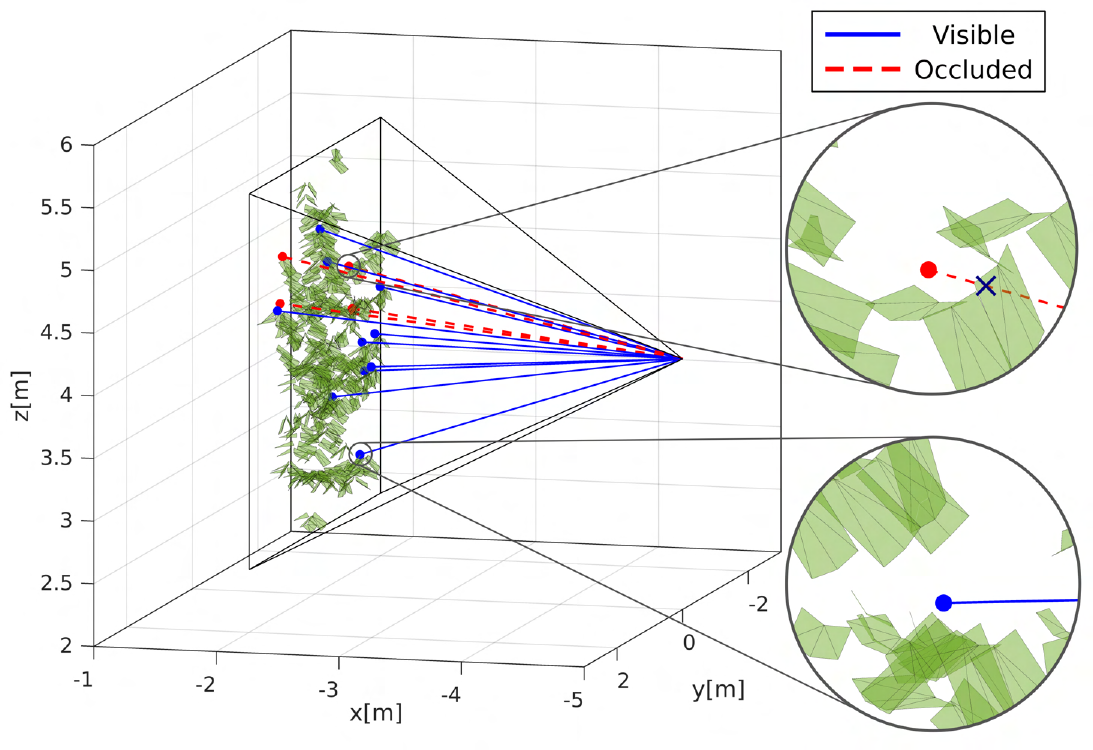}
    \caption{Illustration of the occlusion checking process. Frustum culling is first performed, where only the geometric primitives contained in the camera's field of view are considered. A ray is cast from the camera to each contained fruit's center, and a fruit is marked as visible only if the ray does not intersect any surrounding mesh triangles.}
    \label{Fig:occlusion}
\end{figure}
\section{Orchard Simulation for Trajectory Generation}
\label{Sec:Sim}

\begin{algorithm}[h]
\caption{Visible Fruit Counting Algorithm}
\begin{algorithmic}[1]
\Require $C$: Set of camera configurations
\Require $T$: Set of trees
\Require $F$: Set of fruits on tree $T_j$
\Ensure Detected fruits for each camera configuration $C_i$
\For{$C_i \in C$}
    \State Compute camera frustum $\mathcal{F}(C_i)$
    \For{$T_j \in T$ and $F_k \in F$}
            \If {$F_k$ is inside $\mathcal{F}(C_i)$ }
                \State Cast ray $\mathbf{r}_{C_i \rightarrow F_k}$
                \For{each triangle $\Delta \in T_j$ (leaves, branches, trunks)}
                    \State Check for ray-triangle intersection \cite{moller2005mtalgo}
                    \State \textbf{if} no intersection \textbf{then}
                        \State $V \gets \{T_j, F_k\}$
                \EndFor
            \EndIf
    \EndFor
\EndFor
\State $n_{\text{visible}} \gets |\text{set}(V)|$
\end{algorithmic}
\label{Algo:fruit}
\end{algorithm}

\begin{figure*}[h]
    \centering
    \includegraphics[width=\linewidth]{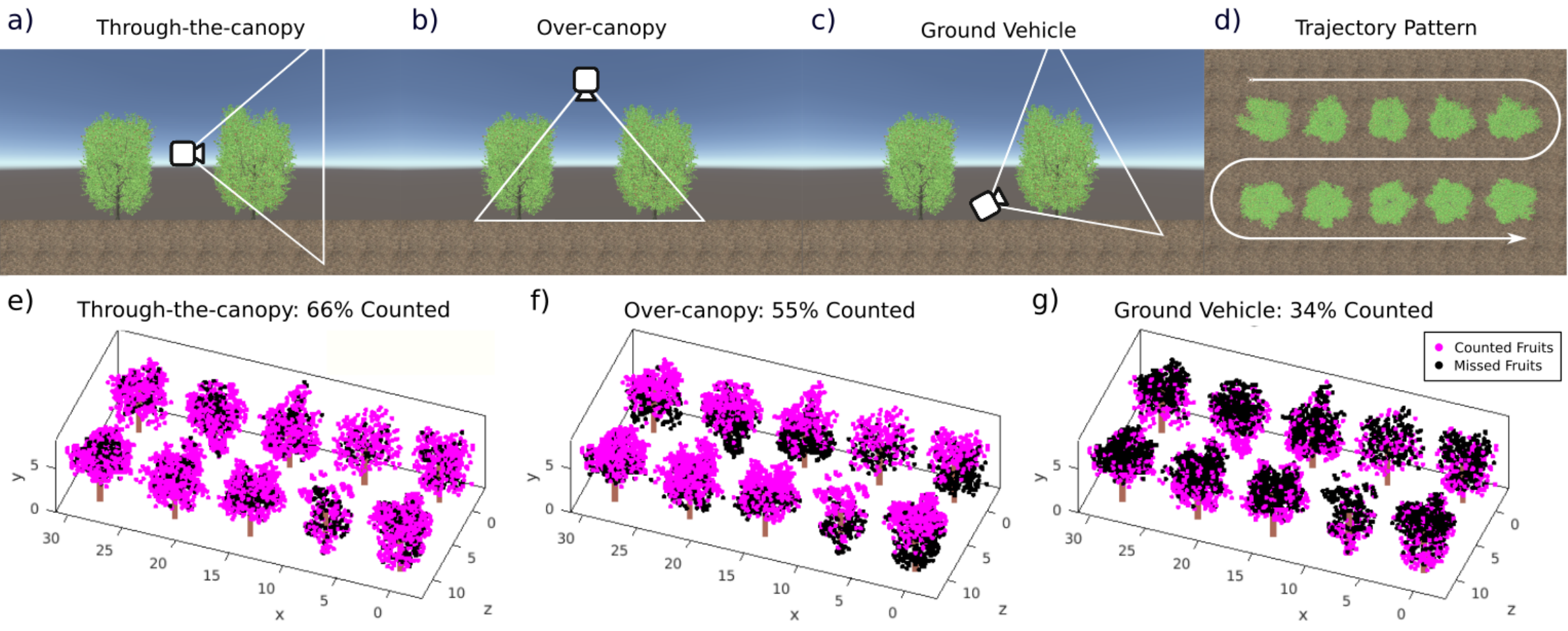}
    \caption{Simulation results comparing fruit counts using three distinct data collection strategies. (a)-(c) depict camera configurations for through-the-canopy flight, over-canopy flight, and ground vehicle collection, respectively. (d) shows the lawn-mower pattern applied uniformly across all methods for comparison. (e)-(g) present fruit counting results and the corresponding fruit distribution patterns. Data collection using the lawn-mower pattern in the simulation captured 66\%, 55\%, and 34\% of the total fruits for each method, respectively, highlighting the advantage of through-the-canopy data collection using UAVs.}
    \label{Fig:three_traj_comp}
\end{figure*}

In this section, we introduce the proposed simulation framework, which provides insights into optimal global trajectory parameters for data collection. A key element of this framework is the realistic modeling of fruit trees and their architectural structures, as accurate representation of the spatial relationships between fruits and tree components is essential for determining fruit visibility. To achieve this, we use the Helios 3D plant modeling framework \cite{bailey2019helios} and its Weber-Penn \cite{weber1995creation} tree generation plug-in, which ensures realistic and adaptable tree architecture generation.

Helios includes models for commonly seen fruit trees, such as orange, walnut, almond, and apple, with built-in parameters such as recursive branching levels, leaf angles, and fruit occurrence that mimic real-world characteristics. Additionally, the canopy generator plug-in allows users to input custom parameters such as tree spacing, trunk height, and fruit radius, which can be adjusted to match the conditions of specific orchards. Since real orchards exhibit varying tree types and spacing, these customizations are crucial to creating a simulation environment tailored to specific properties of the target orchard where autonomous fruit counting will be performed.

\begin{figure}
    \centering
    \includegraphics[width=\linewidth]{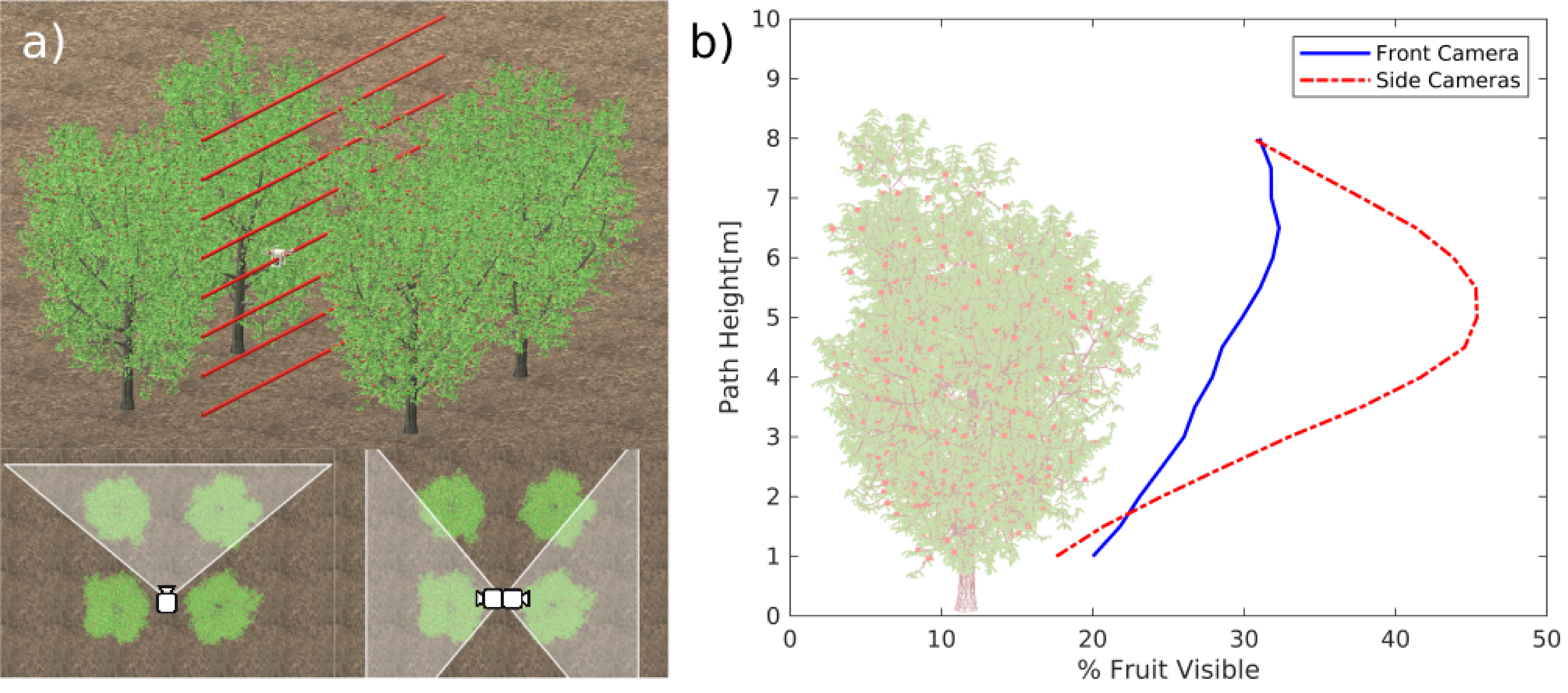}
    \caption{Simulation results comparing fruit counting performance at different through-the-canopy flight heights and camera configurations. (a) Simulated trajectories tested at heights ranging from 1 m to 8 m, using two configurations: a single front-facing camera and two side-facing cameras. (b) Results showing fruit visibility at different flight heights, with optimal heights of 6.5 m for the front-facing camera and 5 m for the side-facing cameras. The two side-facing cameras achieve a maximum visible fruit coverage of 45.4\%, compared to 32.3\% for the front-facing camera.}
    \label{Fig:traj_comp}
\end{figure}

In this simulation environment, all generated trunks, branches, leaves, and fruits are constructed using triangular meshes. Figure~\ref{Fig:occlusion} illustrates selected details focusing on the leaves, while Figure~\ref{Fig:traj_comp}a provides examples of full tree renderings. During this generation process, the mesh vertices are stored, with each tree assigned a unique ID and each fruit assigned a corresponding fruit ID. With this information about the simulated orchard environment, we can perform fruit visibility analysis for a given camera orientation. We assume that branches, leaves, and trunks may obscure the fruits. Given a path with discrete camera orientations and parameters such as field of view and depth of view, the goal is to determine the total number of visible fruits along the path. This process is presented in Algorithm~\ref{Algo:fruit} and a visualization of the occlusion checking process is shown in Figure~\ref{Fig:occlusion}.

The flight trajectories are defined by a set of camera configurations, $C$. For each camera configuration, $C_i$, we can construct a frustum $\mathcal{F}(C_i)$ with an associated field of view and depth that simulates the maximum distance at which a fruit may still be captured by the camera’s resolution. For each tree $T_j$ from the set of generated trees $T$, there is an associated set of fruits denoted as $F$. Frustum culling is first performed, ensuring that only fruits within the camera’s view are checked for occlusion. For each $F_k$ that is contained, a ray $\mathbf{r}_{C_i \rightarrow F_k}$ is cast from the camera location toward the fruit’s center. We then iterate through the mesh triangles of the surrounding geometric primitives, using the Möller–Trumbore algorithm \cite{moller2005mtalgo} to check for intersections. If the ray intersects any triangle between the fruit and the camera, the fruit is considered occluded. Figure~\ref{Fig:occlusion} illustrates this process with examples showing how occlusions are determined by intersections with surrounding geometry. A fruit that passes this occlusion check is marked as visible, and its corresponding fruit ID, along with the associated tree ID, is stored in a multiset $V$. While storing the fruit and tree IDs is not strictly required for counting, doing so provides additional insights into the spatial distribution of visible fruits and their associations with specific trees. This enriched data makes our method valuable not only for fruit counting but also for other applications that benefit from through-the-canopy data collection, such as identifying low-productivity zones or detecting diseases in trees \cite{zhang2021management}.

The proposed simulation framework provides insights into the effectiveness of different methods for collecting fruit counting data. Figure~\ref{Fig:three_traj_comp} illustrates a simulated orchard with 10 walnut trees organized in 10 m by 5 m blocks, which is used to compare three distinct data collection methods, all following a lawn-mower pattern. The simulation includes three camera configurations: a side-mounted camera on a vehicle flying through the canopy, a downward-facing camera on a UAV flying above the canopy, and an upward-facing camera angled at 30 degrees from a 1-meter-tall ground vehicle. The results suggest that the through-the-canopy flight achieves the highest fruit visibility, capturing 66\% of the total fruits visible, compared to 55\% for the over-canopy flight and 36\% for the ground vehicle, emphasizing the need for canopy-level autonomous data collection for this application. By recovering the positions of the fruits through their recorded fruit IDs, this analysis also reveals the distribution of visible fruits for each method, as shown in purple in Figure~\ref{Fig:three_traj_comp}. In practice, such insights can help users optimize flight paths and data collection strategies before deploying vehicles, reducing experimental costs and the need for multiple iterations.

Building on the through-the-canopy flight strategy, we apply the same occlusion model to optimize global trajectory parameters and camera orientations. In Figure~\ref{Fig:traj_comp}, we present an example where a 7.6 m by 7.3 m orchard block is repeatedly generated. In each simulation, the vehicle follows a straight-line path down the center of the 7.6 m spacing, and a range of flight heights from 1 m to 8 m is tested. Two different camera configurations are also compared: one front-facing camera and two side-facing cameras. The fruit counting results for each trajectory and configuration are computed and averaged across the 10 generated orchard blocks. The results indicate that the optimal path height with front-facing camera is 6.5 m, achieving a visual coverage of 32.3\% of all the fruits in the orchard block. In contrast, carrying two side-facing cameras provides significantly greater coverage, reaching 45.4\% at the optimal height of 5 m. It should be noted that the chosen straight-line trajectories inevitably miss fruits on the opposite sides of the trees, which are not covered by the UAV’s flight path, resulting in lower overall percentage coverage compared to the previous example. Additionally, the two side-facing cameras consistently result in higher fruit visibility across nearly all through-the-canopy flight heights, making this configuration a more effective solution for maximizing coverage. Our proposed framework enables users to fine-tune flight paths and camera configurations in simulation, reducing the cost of experimental fine-tuning.

Additionally, the tools introduced in this section can be applied to generate ground truth annotations for synthetic visual data, facilitating the development and testing of vision algorithms, such as those outlined in Section~\ref{Sec:Counting}. By leveraging the known fruit IDs, the 3D positions of the fruits can be accurately projected into the image frame, enabling the creation of annotated datasets. When paired with synthetic images generated by UAV simulators with integrated autonomy stacks, this approach supports comprehensive evaluation of the entire autonomous system, from path planning to data processing. For instance, these capabilities are supported by our previous work on simulating autonomous flight systems in agricultural environments \cite{zha2024agrifly}.

\section{Drone autonomy for data collection}
\label{Sec:Auto}

In this section, we introduce the proposed autonomy pipeline and the vehicle developed for through-the-canopy data collection. The vehicle’s intelligence consists of three key components: a low-level controller that converts thrust and angular velocity commands into motor speeds; the RAPPIDS planner \cite{bucki2020rappids}, which uses depth images for obstacle avoidance during flight; and a visual-inertial-odometry (VIO)-based estimator using OpenVINS \cite{geneva2020openvins}, which integrates IMU and stereo camera inputs for accurate state estimation. The system interfaces through the Robot Operating System (ROS), and a block diagram illustrating these communications is shown in Figure~\ref{Fig:BlockDiagram}.

\begin{figure}
    \centering
    \includegraphics[width=.9\linewidth]{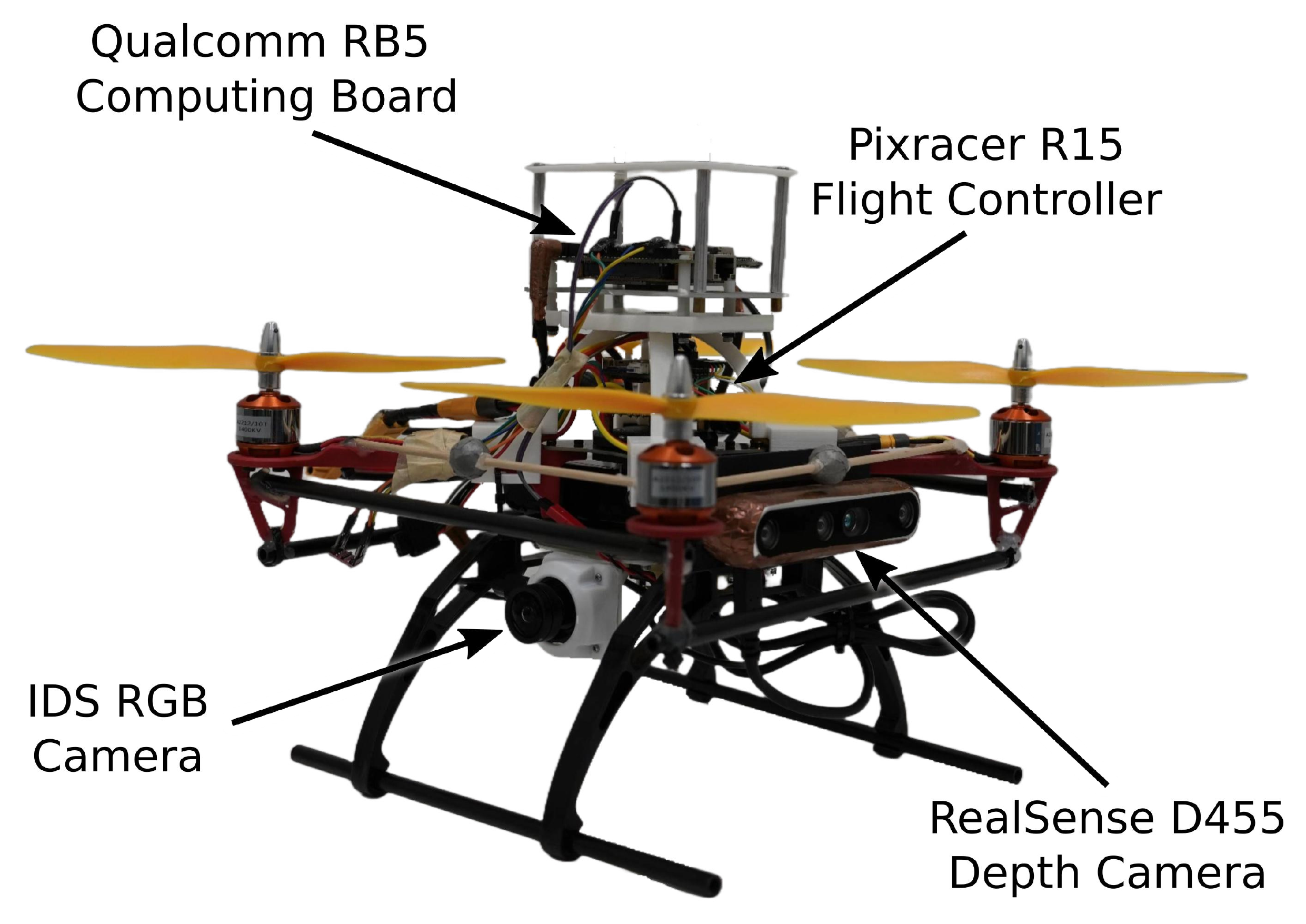}
    \caption{Experimental vehicle with the proposed autonomy stack. A RealSense D455 depth camera captures both depth and stereo images, while an IDS camera is responsible for RGB image collection for fruit counting. The autonomy algorithms are executed on a Qualcomm RB5 board, which sends desired thrust and angular velocity commands to a Pixracer R15 flight controller.}
    \label{Fig:hardware}
\end{figure}

The experimental vehicle comprises of the following components: a RealSense D455 camera for depth and stereo image collection, a Pixracer R15 flight controller, and a Qualcomm RB5 computing board. Additionally, an IDS camera for collecting RGB images is mounted on the side of the vehicle. These hardware components, along with the vehicle design, are depicted in Figure~\ref{Fig:hardware}. Since both state estimation and local planning rely solely on the outputs from the depth camera unit, this design eliminates the need for additional onboard sensors, ensuring cost and power efficiency, and making the vehicle applicable to large-scale monitoring missions. Furthermore, the reliance on vision-based state estimation allows the vehicle to operate in diverse environments, including indoor plant monitoring missions where GPS signals are unavailable \cite{aslan2022indoor}.

\begin{figure}
    \centering
    \includegraphics[width=\linewidth]{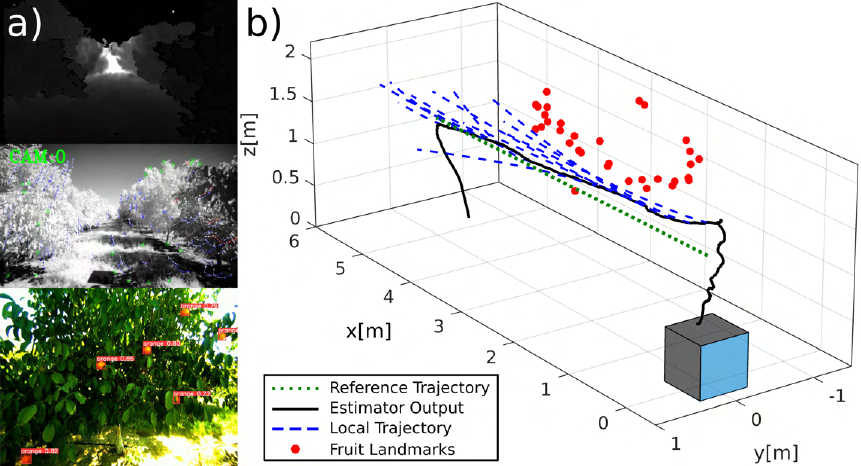}
    \caption{Results from example data collection flight. (a) Samples of the onboard depth image, left stereo image, and RGB image. (b) Visualization of the experimental flight. The supplementary video showcases the same experiment, where the vehicle follows the global reference trajectory (green), and locally planned trajectories (blue) are shown at selected time steps. The VIO-based estimator output is displayed in black, with reconstructed fruit landmarks highlighted in red.}
    \label{Fig:planner}
\end{figure}

Visual-inertial odometry is performed using the OpenVINS framework, which we employ for its accuracy and compatibility, enabling seamless integration with our hardware components. It is based on the Multi-State Constraint Kalman Filter (MSCKF) \cite{mourikis2007msckf}, which efficiently fuses visual and inertial data to estimate the vehicle’s state. The MSCKF method takes inputs from IMU measurements and visual features extracted from stereo images, with the IMU providing measurements at 400 Hz and the stereo images processed at 15 Hz. An example of the left stereo camera image with feature tracking is shown in Figure~\ref{Fig:planner}a. As a result, the vehicle’s 6-degrees-of-freedom pose, along with its velocities and angular velocities, is estimated in real time. 

In addition to accurate state estimation, obstacle avoidance is crucial for ensuring the safety of the robot during through-the-canopy flight. To achieve this, we implement the RAPPIDS local planner \cite{bucki2020rappids}, which uses depth images collected at 15 Hz from the RealSense camera to plan collision-free trajecotries. This feature allows the vehicle to avoid unexpected obstacles, such as tree branches, while in flight. The RAPPIDS planner receives a goal point and conducts local trajectory planning by collision-checking samples of feasible trajectories. Using the simulation framework outlined in Section~\ref{Sec:Sim}, the planned global trajectory is first provided to the planner, and the local planner continuously modifies this path based on depth data during flight. The planner samples minimum jerk trajectories, then checks each sampled trajectory for input feasibility, speed limits, and collisions. The detailed process for feasibility and velocity checking is discussed in \cite{mueller2015jerk}.
For collision checking, the planner partitions the free space into rectangular pyramids based on the depth image, and a trajectory remains collision-free if it stays within the union of these pyramids. This method efficiently reduces the computational cost while maintaining reliable obstacle avoidance. More details about this pyramid generation and collision checking process can be found in \cite{bucki2020rappids}. The planning process is repeated with each new depth image received to account for the latest obstacle information, and the vehicle follows the previously planned trajectory if no new feasible, collision-free path is found.

Finally, we demonstrate the feasibility of our proposed autonomy stack and vehicle design through an experimental flight in an orchard.
The flight is shown in the supplementary video, and Figure~\ref{Fig:FlightSequence} displays an action sequence with selected states of the vehicle. In addition, Figure~\ref{Fig:planner} includes samples of the onboard images and the visualization of local planning results. Specifically, Figure~\ref{Fig:planner}a shows a sample of the onboard depth image, a monochrome image from the stereo camera for visual-inertial odometry, and a collected RGB image with fruits detected in post processing. Figure~\ref{Fig:planner}b presents the reference trajectory in green, the planned trajectory at selected time steps in blue, and the resulting flight path, based on the position output from the estimator, in black. Additionally, using the collected onboard data, we can reconstruct the locations of detected fruits, as outlined in Section~\ref{Sec:Counting}. This example highlights the potential of our proposed data collection method for fruit counting missions.

\section{Autonomous Fruit Counting}
\label{Sec:Counting}

While our proposed methods outlined in previous sections ensure effective data collection flights, accurate fruit identification and counting from the onboard images remain key to completing the autonomous fruit counting mission. To this end, we adopt the approach proposed in \cite{liu2019monocular}, which effectively mitigates double counting, where the same fruit may appear in multiple frames. This method uses fruits as features to reconstruct 3D landmarks, providing an accurate fruit count while maintaining computational efficiency in the reconstruction process. We further enhance this approach by integrating the latest computer vision advancements for detection and tracking. Specifically, we use YOLOv8 \cite{redmon2016yolo} for fruit detection and ByteTrack \cite{zhang2022bytetrack} for tracking across image sequences. The proposed workflow involves: i) detecting fruits in each frame using fine-tuned YOLOv8, ii) tracking detected fruits across the image sequence with ByteTrack, iii) recovering 3D fruit locations via COLMAP \cite{schoenberger2016colmap, schoenberger2016colmap2}, and iv) clustering the reconstructed fruit landmarks to obtain the final count.

\begin{figure}
    \centering
    \includegraphics[width=\linewidth]{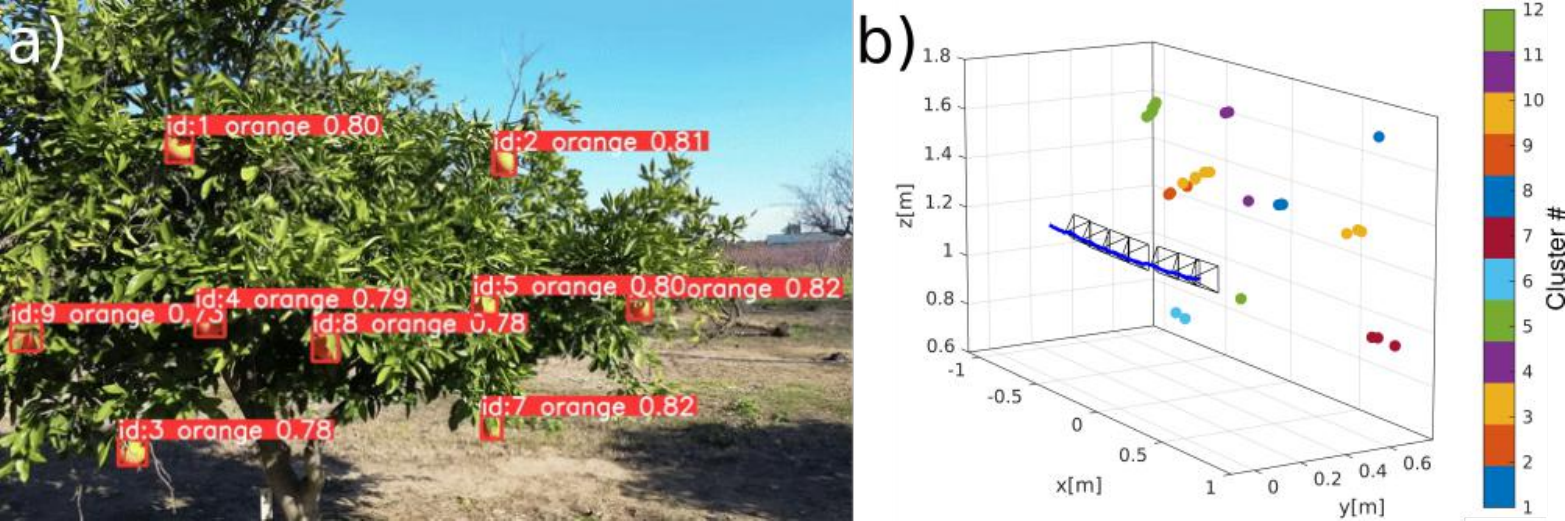}
    \caption{Example results for the proposed autonomous counting workflow. (a) Tracking output where an ID is assigned to each detected and tracked fruit. (b) Counting results after the structure from motion step for a tree with 12 fruits. The 33 reconstructed 3D landmarks are grouped into 12 distinct clusters, denoted by different colors. The reconstructed camera extrinsics are marked in black at selected time steps.}
    \label{Fig:count}
\end{figure}

YOLOv8 is well suited for orchard fruit detection due to its ability to detect multiple overlapping fruits in cluttered natural environments while maintaining high accuracy even for smaller, partially obscured objects. We fine-tune the YOLOv8 model using a dataset of 100 manually labeled images of the target fruit type, collected under good lighting conditions. This process enables the model to adapt to the unique visual characteristics of the fruit, ensuring robust performance even under suboptimal conditions, such as variations in fruit color caused by lighting changes. In the video shown in Figure~\ref{Fig:count}, the YOLOv8 model, without fine-tuning, detects 72.7\% of all fruit occurances with an average confidence of 39.6\%, while only 52.4\% of the detected fruits are correctly classified as oranges, with the rest classified as apples.  After fine-tuning with images captured in similar conditions, the model detects 100\% of the fruits with an average confidence of 79.6\% and correctly classifies them as oranges.

While \cite{liu2019monocular} uses the Hungarian Algorithm for fruit tracking, we adopt ByteTrack for its superior handling of low-confidence detections, which are common under poor imaging conditions. ByteTrack effectively maintains feature correspondences between frames, providing robust raw inputs for the structure-from-motion step. Figure~\ref{Fig:count}a shows a sample of the tracking output, with each tracked fruit assigned a unique ID to distinguish it through the sequence.

COLMAP remains the state-of-the-art structure-from-motion toolbox, and we use it to reconstruct the 3D landmarks representing the fruits. Using the feature correspondences from the tracking step, COLMAP recovers camera intrinsics, extrinsics, and 3D landmarks, with fruits serving as features. Since the RGB camera is integrated into the autonomy stack, the reconstruction process may be initialized using the camera extrinsics, improving the reliability and accuracy of fruit landmark positioning. Finally, we apply density-based spatial clustering \cite{ester1996dbscan} to group the reconstructed landmarks by individual fruits, yielding the final fruit count.

Figure~\ref{Fig:count} illustrates an example of autonomous fruit counting using the proposed workflow. A video captured during a low-altitude flight near a small orange tree with 12 fruits is processed, and the method accurately recovers all 12 fruits along with their estimated positions. The fruits are first detected and tracked in the image frames, as shown in Figure~\ref{Fig:count}a. During the structure from motion step, the camera extrinsics and 33 landmarks are reconstructed, as depicted in Figure~\ref{Fig:count}b. Finally, after applying the clustering algorithm, the landmarks are grouped into 12 distinct clusters, as shown in different colors, revealing the total fruit count. This example with orange counting demonstrates the effectiveness of the proposed approach. However, in larger orchards with smaller fruits, increased occlusions may impact detection accuracy, as discussed in Section~\ref{Sec:Sim}. With further refinement, the pipeline is expected to handle these more challenging scenarios effectively.

\section{Conclusion and Future Work}
In this work, we have developed an autonomous aerial system designed for efficient and safe through-the-canopy fruit counting in large-scale orchards. Our approach leverages a high-fidelity simulation framework to optimize global flight trajectories, a cost-effective autonomy stack for canopy-level navigation and data collection, and state-of-the-art computer vision tools for fruit counting. The system's performance was validated through successful fruit counting using aerial images captured at canopy level, along with demonstrations of the autonomous navigation capabilities of the experimental vehicle. These results demonstrate the system's potential to enhance data collection efficiency and fruit yield estimation in complex orchard environments, offering a promising solution for large-scale agricultural applications.

For future work, several improvements can be made to further enhance the system's capabilities. First, global planning could be improved by considering the flight speed of the vehicle, as it significantly impacts both power consumption and orchard coverage. Second, incorporating real-time feedback on the quality of the collected images could further ensure the reliability of data collection. While the integration of the RGB camera into the autonomy stack provides the potential for this enhancement, it has not yet been implemented in the current system. Finally, closer integration of the autonomy stack and fruit counting pipeline would enable real-time, onboard fruit counting, which is achievable given the computational efficiency of our method. Addressing these factors would help further optimize the system for large-scale, continuous agricultural monitoring.

\section*{ACKNOWLEDGMENT}
This work was supported by the Agriculture and Food Research Initiative (AFRI) Competitive Grant no. 2020-67021-32855/project accession no. 1024262 from the USDA National Institute of Food and Agriculture. The authors would like to thank Dylan Lee and Ting-Hao Wang for their contributions to the development of the autonomous vehicle, and Brian Bailey along with the members of the UC Davis Plant Simulation Laboratory for their valuable advice on plant modeling and fruit detection.
\bibliographystyle{IEEEtran}
\bibliography{references.bib}

\end{document}